\pdfoutput=1

\documentclass[11pt]{article}

\usepackage[final]{acl}

\usepackage{times}
\usepackage{latexsym}

\usepackage[T1]{fontenc}

\usepackage[utf8]{inputenc}

\usepackage{microtype}

\usepackage{inconsolata}

\usepackage{graphicx}
\usepackage{booktabs}
\usepackage{multirow}
\usepackage{amsmath}
\usepackage{colortbl}
\usepackage{xcolor}
\usepackage{enumitem}
\usepackage{amsfonts}

\title{LADM: Long-context Training Data Selection with Attention-based Dependency Measurement for LLMs}

\author{
    Jianghao Chen\textsuperscript{1,2,3}, \ 
    Junhong Wu\textsuperscript{1,2},\ 
    Yangyifan Xu\textsuperscript{1,2}, \
    Jiajun Zhang\textsuperscript{1,2,4}\thanks{\ \ Corresponding Author}   \\
    \textsuperscript{1}Institute of Automation, Chinese Academy of Sciences\\
    \textsuperscript{2}School of Artificial Intelligence, University of Chinese Academy of Sciences\\
    \textsuperscript{3}Zhongguancun Academy, Beijing, China \textsuperscript{4}Wuhan AI Research\\
    \texttt{\{chenjianghao2022, wujunhong2021, xuyangyifan2021\}@ia.ac.cn}\\
    \texttt{jjzhang@nlpr.ia.ac.cn} \\
}

\begin{document}
\maketitle
\begin{abstract}
Long-context modeling has drawn more and more attention in the area of Large Language Models (LLMs). Continual training with long-context data becomes the de-facto method to equip LLMs with the ability to process long inputs. However, it still remains an open challenge to measure the quality of long-context training data. To address this issue, we propose a \textbf{L}ong-context data selection framework with \textbf{A}ttention-based \textbf{D}ependency \textbf{M}easurement (\textbf{LADM}), which can efficiently identify high-quality long-context data from a large-scale, multi-domain pre-training corpus. LADM leverages the retrieval capabilities of the attention mechanism to capture contextual dependencies, ensuring a comprehensive quality measurement of long-context data. Experimental results show that our LADM framework significantly boosts the performance of LLMs on multiple long-context tasks with only 1B tokens for continual training. \footnote{Our code is available at \url{https://github.com/ZNLP/LADM}}
\end{abstract}

\section{Introduction}
Long-context modeling for Large Language Models (LLMs) has recently drawn more and more attention. The maximum context window of current LLMs has been significantly extended to 128K tokens for GPT-4 \citep{achiam2023gpt} and 1M tokens for Gemini 1.5 \citep{reid2024gemini}. The long-context modeling ability greatly contributes to more sophisticated applications of LLMs in various fields such as long-context retrieval \citep{needleinhaystack, xu2023retrieval, an2024make}, question answering \citep{kovcisky2018narrativeqa, dasigi2021dataset, bai2023longbench}, and summarization \citep{huang2021efficient, zhong2021qmsum, fabbri2019multi}.

Preparing long-context dataset and performing continual training has become the de-facto framework to enrich LLMs with the ability to process long inputs. However, quality measurement of the long-context training data has not received enough attention. Previous studies show that if the training data is composed of concatenated short samples or lacking dependencies over long contexts \citep{ding2024fewer, fu2024data, chen-etal-2024-long}, models may fail to learn how to handle long-range and diverse contextual dependencies. These low-quality training samples can aggravate the tendency of LLMs to ignore valuable distant contextual information, further limiting their performance on long-context tasks. Therefore, measuring the quality of long-context training data and mining high-quality training samples become crucial for enhancing the long-context modeling capability of LLMs.

The dependency between segments within context is the key indicator of high-quality long-context data. Recent studies have proposed several methods to take long-context dependencies into consideration when constructing pre-training data. \citet{staniszewski2023structured} enhance long-range semantic dependencies by integrating relevant documents into one sample. \citet{chen-etal-2024-long} split long-context data into segments and measure dependencies through delta perplexity scores between individual segments. However, these methods ignore the inherent structures and relationships within long contexts, leading to inaccurate measurement of long-range contextual dependencies.

To address the above issue, we propose a \textbf{L}ong-context data selection framework with \textbf{A}ttention-based \textbf{D}ependency \textbf{M}easurement (\textbf{LADM}), which measures the long-context dependency with span-level attention scores. Inspired by the inherent retrieval operations of the attention mechanism \citep{mittal2022compositional, wu2024retrieval}, LADM aims to leverage the attention distribution to measure the relationship within long contexts. Specifically, we first train a tiny model with long-context modeling capability named Long Attention Calculator. Then, to measure the dependency of a single long-context sample, we feed the sample into the Long Attention Calculator and compute the \textbf{P}airwise \textbf{F}ocus \textbf{S}core (\textbf{PFS}) between different spans by the accumulated attention scores. Subsequently, the \textbf{A}ggregated \textbf{F}ocus \textbf{S}core (\textbf{AFS}) for each span is derived by incorporating all PFS between this span and its preceding ones. Finally, we define the sample-level \textbf{C}ontextual \textbf{D}ependency \textbf{S}core (\textbf{CDS}) by a weighted sum of all AFS and select samples with high CDS for continual pre-training. Experimental results show that LADM outperforms other data selection methods on various long-context tasks, achieving an average performance improvement of 2.16\% for four models across different sizes and architectures on the LongBench dataset.

Our contributions are summarized as follows: 1) We propose an efficient data selection framework LADM, which can identify high-quality long-context data with long-range and diverse contextual dependencies from a large-scale pre-training corpus. 2) We introduce a novel method for dependency measurement through the attention mechanism, effectively capturing long-range and diverse relationships within the complete contextual information. 3) Experimental results demonstrate the superiority of our method. We achieve better performance with only half of the pre-training tokens than the random sampling approach.

\section{Related Work}
\subsection{Long-context Modeling for LLMs}
Enhanced long-context modeling in LLMs can drive substantial progress in artificial intelligence across various domains, such as long-chain reasoning \citep{chen2025lr, yeo2025demystifying, sun2025efficient, sun2025ktae}, long video and image processing \citep{weng2024longvlm, zhang2024simple, guan-etal-2025-trifine, jian2024large}, and long-form generation \citep{chen2025longpo, bai2024longwriter, chen2025ace}.
To enable better long-context processing capability of LLMs, recent studies have explored both training-free and training-augmented methods. For training-free methods, \citet{xiao2024efficient} and \citet{han2024lm} focus on retaining the attention on the initial and local tokens while masking those at greater distances, thereby enhancing the length generalization ability of LLMs. DCA \citep{an2024trainingfree} and SelfExtend \citep{jin2024llm} rearrange position indices of long-context inputs and get impressive length extrapolation capability without fine-tuning. Training-augmented methods involve continuing to pre-train LLMs on longer contexts with modified positional encoding. Positional Interpolation (PI) \citep{chen2023extending}, NTK \citep{ntkaware}, and YaRN \citep{peng2024yarn} effectively achieve context window extension through the interpolation and extrapolation of RoPE positional embedding \citep{su2024roformer}. Moreover, LongLoRA \citep{chen2024longlora} combines PI with $\text{S}^2\text{-Attn}$ and LoRA \citep{hu2021lora}, enabling more efficient training. \citet{zhu2024coca} ensure a collinear constraint between query and key vectors when integrating RoPE into self-attention and pre-train LLMs with better long-context extrapolation ability. To fully leverage these methods, it is essential to construct high-quality long-context training data.

\subsection{Long-context Training Data for LLMs}
Training data quality is crucial for the long-context modeling capability of LLMs. \citet{fu2024data} focuses on domain balance and length up-sampling for long-context training data. \citet{staniszewski2023structured} and \citet{gao2024quest} propose similarity-based approaches of grouping documents to construct long-context training data. \citet{he-etal-2024-never} designs synthesized multi-doc QA tasks and improves the long-context information searching and reflection ability. To alleviate the "lost-in-the-middle" \citep{liu2024lost} phenomenon, \citet{an2024make} introduces Information-Intensive training on long-context QA tasks and \citet{xiong2024artificial} designs key-value retrieval tasks for fine-tuning.

Our approach characterizes the contextual dependencies within long-context samples based on the attention mechanism and effectively identifies high-quality long-context data from a large pre-training corpus. Concurrently with our work, \citet{chen-etal-2024-long} proposes a framework ProLong that divides a long-context sample into segments and calculates the delta perplexity scores between individual segments as Dependency Strength without the original context. However, we capture dependencies within the complete context, providing a more comprehensive view of the contextual relationships. Moreover, our experiments are conducted on a more diverse pre-training corpus, proving the robustness and applicability of our method.

\section{Method}

\begin{figure*}[t]
  \includegraphics[width=\linewidth]{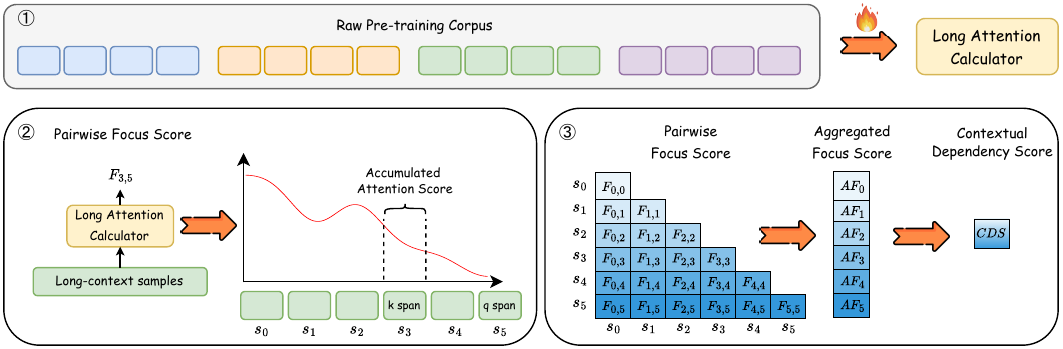}
  \caption{The overall framework of LADM. We first train a Long Attention Calculator, then calculate the Pairwise Focus Score (PFS) to measure the dependency between spans. Then, we compute the Aggregated Focus Score (AFS) of each span and merge them as the Contextual Dependency Score (CDS) of a single long-context sample.}
  \label{framework}
  \vspace{-4mm}
\end{figure*}

\subsection{Problem Formulation}
The long-context modeling capability of LLMs is obtained by continual pre-training on long-context samples. Given a dataset $\mathcal{D}$ of long-context samples with different quality levels, we aim to select a high-quality subset $\mathcal{D}_s$. Specifically, we define a scoring function $s$ to measure the quality of each data sample $x \in \mathcal{D}$. Then we rank all the samples according to their scores $s(x)$ and select the top $N$ samples to construct the subset $\mathcal{D}_s$ as follows:
\begin{equation}\label{selection}
    \mathcal{D}_s = \{ x \in \mathcal{D} \mid \text{rank}(s(x)) \leq N \}
\end{equation}

\subsection{Preliminary Analysis}
\setlength{\tabcolsep}{4pt}
\begin{table}[htbp]
    \centering
    {\renewcommand{\arraystretch}{0.9}\resizebox{1.0\columnwidth}{!}{\begin{tabular}{cccccccccc}
        \toprule
        \textbf{Data} & \multicolumn{9}{c}{\textbf{Evaluation Context length}} \\
        \textbf{Length} & 4K & 8K & 12K & 16K & 20K & 24K & 28K & 32K & \textbf{Avg} \\
        \midrule
        4K & \cellcolor{green!50}1.00 & \cellcolor{yellow!50}0.80 & \cellcolor{green!40}0.90 & \cellcolor{yellow!40}0.76 & \cellcolor{red!50}0.22 & \cellcolor{red!30}0.40 & \cellcolor{red!50}0.20 & \cellcolor{red!40}0.30 & 0.57 \\
        8K & \cellcolor{green!50}1.00 & \cellcolor{yellow!25}0.60 & \cellcolor{yellow!50}0.80 & \cellcolor{green!50}1.00 & \cellcolor{orange!30}0.53 & \cellcolor{yellow!40}0.66 & \cellcolor{orange!30}0.50 & \cellcolor{red!30}0.43 & 0.69 \\
        16K & \cellcolor{green!50}1.00 & \cellcolor{green!50}1.00 & \cellcolor{yellow!25}0.60 & \cellcolor{green!40}0.90 & \cellcolor{yellow!40}0.70 & \cellcolor{green!50}1.00 & \cellcolor{green!40}0.90 & \cellcolor{yellow!40}0.70 & 0.85 \\
        32K & \cellcolor{green!50}1.00 & \cellcolor{green!40}0.90 & \cellcolor{yellow!50}0.80 & \cellcolor{yellow!25}0.60 & \cellcolor{green!40}0.90 & \cellcolor{green!40}0.90 & \cellcolor{green!40}0.90 & \cellcolor{green!50}1.00 & 0.88 \\
        \bottomrule
    \end{tabular}}}
    \caption{The "Needle-in-the-Haystack" performance on models trained with data concatenated at different lengths. For each evaluation context length, the result is the average of all performances across needle insertion depths, ranging from 0 to 1.}
    \label{pre_exp}
    \vspace{-2mm}
\end{table}

We first conduct a preliminary experiment to analyze the impact of contextual dependencies on LLMs' long-context modeling capability. We train the Llama2-7B model with 32K-token sequences concatenated by 4K$\times$8, 8K$\times$4, and 16K$\times$2-token samples, respectively, and compare them with one trained with samples exceeding 32K. The total number of training tokens amounts to 0.5B.

As shown in Table \ref{pre_exp}, we can observe that the average retrieval accuracy decreases as the original length of training data becomes shorter. This trend indicates that models trained with concatenated and contextually unrelated data may focus more on local information and lack retrieval ability across long contexts. Consequently, training data with strong contextual dependencies is essential for enhancing the long-context modeling capability of LLMs. These findings align with recent studies \citep{staniszewski2023structured, chen-etal-2024-long}.

Building on this analysis, we further explore how to measure the contextual dependencies within long-context data. \citet{wu2024retrieval} reveal that LLMs incorporate retrieval ability within the attention mechanism, which can be reflected in the attention distribution, with higher weights assigned to previous tokens more related to the current token. Inspired by this, we propose LADM framework to quantify the contextual dependencies by analyzing the attention distribution over long contexts.

\subsection{LADM Framework}

Figure \ref{framework} illustrates the overall framework of LADM. We first train a Long Attention Calculator with basic long-context modeling capability and use it to calculate the Pairwise Focus Score (PFS), which measures the dependency between spans within long contexts. Then, the Aggregated Focus Score (AFS) for each span is derived by incorporating all PFS between this span and previous spans. Finally, the sample-level Contextual Dependency Score (CDS) is computed by merging all the AFS of different spans.

\begin{figure}[!t]
  \centering
  \includegraphics[width=0.9\columnwidth]{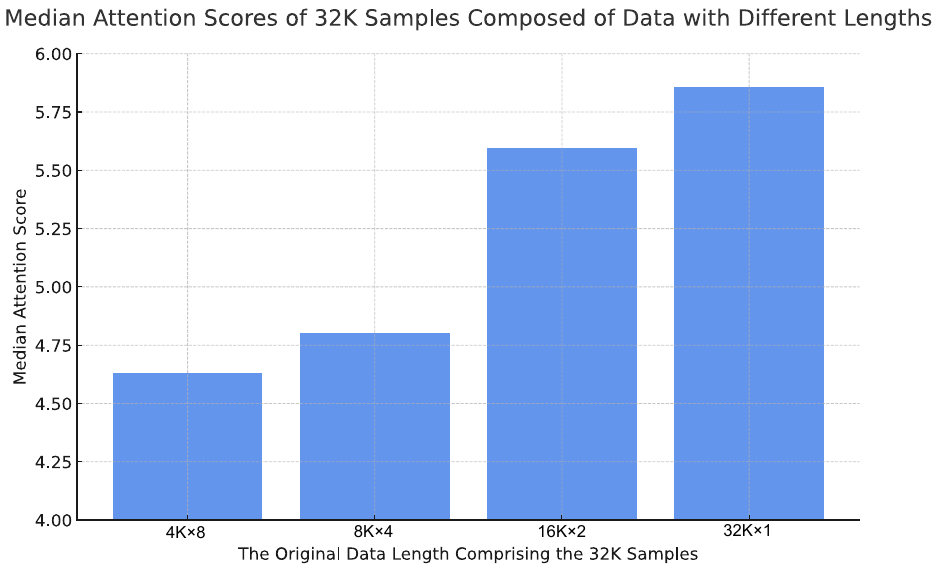}
  \caption{The median attention scores under different 32K data sample construction methods.}
  \label{attn}
  \vspace{-4mm}
\end{figure}

\paragraph{Long Attention Calculator} To leverage the intrinsic retrieval capability of the attention mechanism for detecting dependency within long contexts, we present Long Attention Calculator, a compact model with basic long-context modeling capability. Specifically, we choose the TinyLlama-1.1B-v1.1 \footnote{\url{https://huggingface.co/TinyLlama/TinyLlama_v1.1}} for subsequent data filtering efficiency and train it with randomly sampled 32K-token sequences. To demonstrate the Long Attention Calculator's ability to capture dependencies within long contexts, we use it to calculate the median value of accumulated attention scores between long-range spans on 1,000 32K samples concatenated by data with different lengths. As shown in Figure \ref{attn}, the results indicate that the Long Attention Calculator can distinguish samples with varied dependencies via attention scores over long contexts, as the complete 32K samples exhibit higher attention scores on distant previous tokens than concatenated ones. Therefore, we can design metrics based on attention scores to measure the dependency within long-context data.

\paragraph{Pairwise Focus Score}
Given a long-context sample $S$ consisting of $N$ spans of length $l$:  $S = \{s_0, s_1, ..., s_{N-1}\}$, the Pairwise Focus Score $\mathrm{PFS}(i, j)$ between $s_i$ and $s_j$ (where $j > i$) is defined as follows:
\begin{equation}\label{fs}
    \scalebox{0.95}{$\mathrm{PFS}(i, j)=\mathrm{Sum}\left(\mathrm{Softmax}\left(\frac{Q_{j}K^{T}_{0:j}}{\sqrt{d_k}}\right)[:, i]\right)$}
\end{equation}
where $Q_j$ is the query states of $s_j$, $K_{0:j}$ is the key states of $s_0$ to $s_j$ and $\frac{1}{\sqrt{d_k}}$ is the scaling factor. $\mathrm{PFS}(i, j)$ calculates the accumulated attention weights that $s_j$ assigns to $s_i$, quantifying the influence of $s_i$ on the final representation of $s_j$ in the attention mechanism. Therefore, we can effectively detect spans that exhibit strong dependencies within long contexts by calculating PFS with the Long Attention Calculator.

To comprehensively evaluate the contextual dependencies of a long-context sample, it is essential to aggregate PFS between different spans. This approach can help us gain a deeper understanding of the complex relationships between various parts of the long-context data. We consider the following criteria for aggregating multiple PFS:

\paragraph{Aggregated Focus Score} For each span $s_j$, we calculate PFS at varied intervals, including $\mathrm{PFS}(0, j)$, $\mathrm{PFS}(1, j)$, ..., $\mathrm{PFS}(j-1, j)$. This aids in understanding how the influence of previous spans varies with increasing intervals, revealing both short-range and long-range dependencies. Notably, recent studies \citep{xiao2024efficient, hsieh2024found} have shown that initial and recent tokens often receive disproportionate attention weights, suggesting a predominance of the beginning and local dependencies. Therefore, we exclude scores for the first $m$ spans ($s_0, ..., s_{m-1}$) and the local $n$ spans ($s_{j-n}, ..., s_{j-1}$). We select previous spans at a stride of $d$ for further computational efficiency.

We apply weights to these scores based on the lengths of the intervals, thus encouraging longer-distance dependencies. Additionally, we incorporate the variance of these scores. A higher variance indicates that $s_j$ exhibits more diverse dependencies from its previous context, reflecting a complex dependency pattern and rich structural information within the long-context sample. Therefore, we define the Aggregated Focus Score (AFS) of span $s_j$ as follows (omitting stride sampling for clarity):
\begin{equation}\label{sig}
    \scalebox{0.9}{$\sigma_{j} = \sigma\left(\mathrm{PFS}(m, j), ..., \mathrm{PFS}(j-n-1, j)\right)$}
\end{equation}
\begin{equation}\label{afs}
    \scalebox{0.9}{$\mathrm{AFS}(j) = \sigma_{j} \displaystyle \sum_{i=m}^{j-n-1} \frac{j-i}{N} \cdot \mathrm{PFS}(i, j)$}
\end{equation}
where $\sigma_{j}$ is the standard deviation of all PFS.

\paragraph{Contextual Dependency Score} While $\mathrm{AFS}(j)$ provides a measurement of the dependencies for a single span $s_j$, we should further consider the contributions of all spans to accurately represent the overall dependencies of a long-context sample. To achieve this, we sum all AFS and apply a weight based on index $j$ to highlight the contributions of spans with larger positions. This approach aligns with our focus on long-range dependencies, as spans with larger indices have the potential to consider previous spans at greater distances. The sample-level Contextual Dependency Score (CDS) is therefore defined as follows:
\begin{equation}\label{cds}
    \scalebox{0.9}{$\mathrm{CDS}(S) = \displaystyle \sum_{j=n_0}^{N-1} \frac{j}{N} \cdot \mathrm{AFS}(j)$}
\end{equation}
where $n_0$ is the index of the first span and $d$ is the stride. We exclude the initial spans' $\mathrm{AFS}(j<n_0)$, as these early spans have fewer previous spans to depend on, resulting in a less informative measurement of dependencies. For computational efficiency, we calculate AFS at a stride of $d$ (omitted from Eq.\ref{cds} for clarity).

\section{Experiments Settings}
\subsection{Pre-training Dataset}
We use the Pile \citep{gao2020pile} corpus for long-context continual pre-training. Our experiments are conducted on data samples with 32K tokens due to the scarcity of longer samples. We remove samples with lengths less than 32K measured by LlamaTokenizer. The detailed information of our pre-training dataset is provided in Appendix \ref{appendixA}.

\subsection{Baselines}
We compare LADM with the following methods:
\paragraph{Random Sampling} We randomly sample long-context data from the dataset for continual training.
\paragraph{ProLong} \citet{chen-etal-2024-long} propose a framework filtering long-context data with delta perplexity scores between individual segments. We follow all settings of Prolong except changing the model to TinyLlama-1.1B-v1.1 for a fair comparison.

\subsection{Implementation Details}
\paragraph{Hyper-parameters} When calculating PFS, we truncate all samples to 32K-token sequences for batch operation. The span size is set to $l=128$, resulting in $N=256$ spans per sample. Span-level AFS uses $m=1, n=d=4$, excluding the first and recent four spans and selecting previous spans at a stride of four. Sample-level CDS uses $n_0=16, d=4$, excluding the initial 16 spans' AFS and calculating AFS at a stride of four.

\paragraph{Data Selection} We rank all samples with their CDS and select the top-ranking samples from each data domain, maintaining the original domain distribution. All methods use 1B tokens selected from the dataset for continual training.

\paragraph{Training Configuration} We increase the base of RoPE from 10,000 to 500,000 following \cite{xiong2023effective}. For the Long Attention Calculator, since the original context length of TinyLlama is 2K, we use 5B randomly sampled training tokens to ensure better long-context dependencies measurement. We also provide experimental results using the Long Attention Calculator trained with 1B tokens in Appendix \ref{appendixE} for comparison. We continually pre-train OpenLlama-3B-v2 \citep{openlm2023openllama}, Llama2-7B/13B \citep{touvron2023llama2} and Mistral-7B-v0.1 \citep{jiang2023mistral} with 32K-token sequences for 1B tokens. More training details are displayed in Appendix \ref{appendixB}.

\subsection{Evaluation Tasks and datasets}
We take the following tasks to evaluate the long-context capability of LLMs: 

\begin{figure*}[t]
  \includegraphics[width=0.85\linewidth]{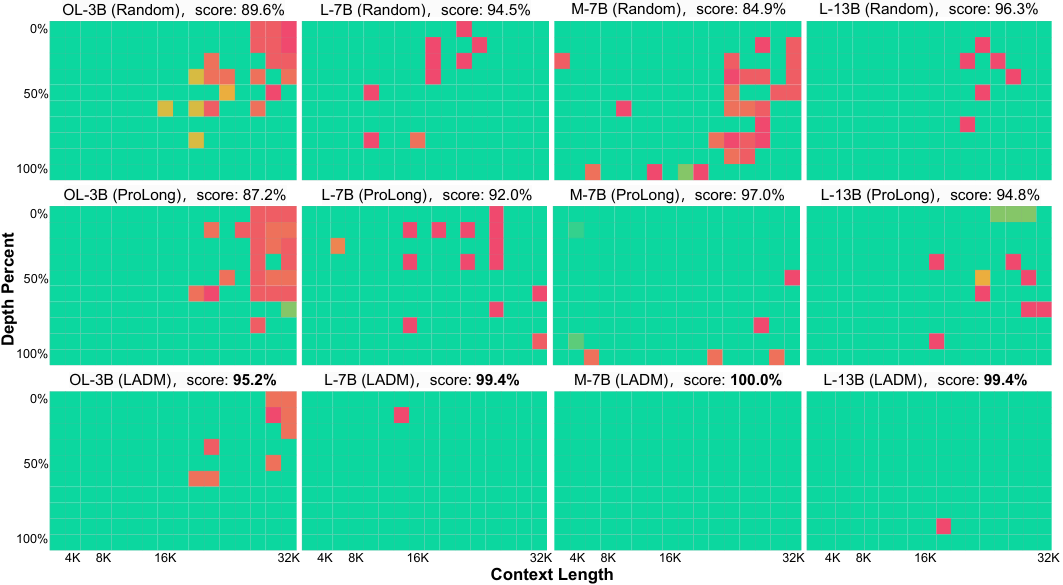}
  \centering
  \caption{The "Needle-in-the-Haystack" task performance of different data selection methods. The x-axis denotes the evaluation context length, and the y-axis denotes insertion depths of the "needle".}
  \label{needle}
  \vspace{-2mm}
\end{figure*}

\paragraph{Perplexity Evaluation}
We evaluate the language modeling capability of LLMs by measuring the perplexity (PPL) on real-world long-context data. We collect samples exceeding 32K from the test split of the Proof-Pile \citep{proofpile} dataset and calculate the average PPL across different context window sizes.

\paragraph{Synthetic Tasks Evaluation}
We test the long-context retrieval ability of LLMs on the "Needle-in-the-Haystack" task \citep{needleinhaystack}. This synthetic task is designed to evaluate LLMs' capability to locate essential information across varying positions and context lengths.

\paragraph{Real-World Tasks Evaluation}
Evaluation on perplexity or synthetic retrieval tasks can not truly reflect the performance of LLMs under real-world scenarios \citep{hu2024can, fang2024wrong}. Therefore, we select different types of tasks from LongBench \citep{bai2023longbench} for further evaluation.

\section{Experimental Results}
\label{section5}
We evaluate LADM with OpenLlama-3B-v2, Llama2-7B/13B and Mistral-7B-v0.1 on language modeling tasks, synthetic long-context tasks and real-world long-context tasks. We use "OL-3B", "L-7B", "L-13B", and "M-7B" to denote these models for short.

\subsection{Perplexity Evaluation}
Table \ref{ppl} shows the perplexity of long-context samples from Proof-Pile across various context window sizes. The perplexity differences among the three methods are minimal due to the same domain distribution of the training data. Our proposed LADM outperforms other methods under all model and context window size settings. This demonstrates the superiority of our models on long-context language modeling tasks.

\subsection{Synthetic Tasks Evaluation}
In Figure \ref{needle}, we compare the retrieval accuracy of our models with other baseline models on the "Needle-in-the-Haystack" task. Notably, our models show higher average retrieval accuracy and even achieve nearly 100\% retrieval rate for Llama2-7B/13B and Mistral-7B with only 1B training tokens. Compared to our method, other methods exhibit inferior performance, particularly in scenarios involving greater retrieval distances or when the "needle" is located in the middle of the context.

\subsection{Real-World Tasks Evaluation}
We report the experimental results on real-world long-context tasks in Table \ref{longbench}. For the randomly sampling method, we conduct three experiments with different random sampling seeds and calculate the average performance. Our LADM framework outperforms other methods across nearly all types of tasks and all models. Specifically, our method achieves an average performance improvement of 2.16\% on four models over ProLong. Notably, for Mistral-7B, LADM achieves a substantial improvement of 10.09\% on single-document QA and 4.66\% on multi-document QA. These results highlight the strong capability of our models in handling real-world tasks and validate the effectiveness of our data selection framework, indicating the importance of incorporating long-range contextual dependencies into training data. Experimental results in Appendix \ref{appendixF} also show that our method enhances LLMs' long-context capabilities while preserving short-context performance.

\begin{table}[t]\normalsize
    \centering
    {\renewcommand{\arraystretch}{1.0}\resizebox{1.0\columnwidth}{!}{\begin{tabular}{llccccc}
        \toprule
        \multirow{2}{*}{\textbf{Model}} & \multirow{2}{*}{\textbf{Method}} & \multicolumn{5}{c}{\textbf{Context Window Size}} \\
         &  & 2K & 4K & 8K & 16K & 32K \\
        \midrule
        \multirow{3}{*}{OL-3B} & Random & 4.951 & 4.268 & 3.562 & 3.026 & 2.685 \\
        & ProLong & 4.941 & 4.271 & 3.573 & 3.040 & 2.701 \\
        & LADM & \textbf{4.910} & \textbf{4.247} & \textbf{3.553} & \textbf{3.022} & \textbf{2.683} \\
        \midrule
        \multirow{3}{*}{L-7B} & Random & 4.515 & 3.900 & 3.264 & 2.780 & 2.458 \\
        & ProLong & 4.516 & 3.906 & 3.275 & 2.792 & 2.470 \\
        & LADM & \textbf{4.481} & \textbf{3.878} & \textbf{3.252} & \textbf{2.773} & \textbf{2.453} \\
        \midrule
        \multirow{3}{*}{M-7B} & Random & 4.620 & 3.936 & 3.267 & 2.775 & 2.455 \\
        & ProLong & 4.293 & 3.696 & 3.095 & 2.644 & 2.346 \\
        & LADM & \textbf{4.266} & \textbf{3.673} & \textbf{3.076} & \textbf{2.629} & \textbf{2.332} \\
        \midrule
        \multirow{3}{*}{L-13B} & Random & 4.200 & 3.657 & 3.084 & 2.645 & 2.349 \\
        & ProLong & 4.200 & 3.659 & 3.090 & 2.651 & 2.356 \\
        & LADM & \textbf{4.176} & \textbf{3.637} & \textbf{3.070} & \textbf{2.636} & \textbf{2.342} \\
        \bottomrule
    \end{tabular}}}
    \caption{\label{ppl}Perplexity of long-context samples from the test split of the Proof-Pile dataset. Each sample is truncated to the corresponding context window size.}
    \vspace{-4mm}
\end{table}

\section{Analysis}

\subsection{Training Efficiency of LADM}
\begin{table*}[t]\normalsize
    \centering
    {\renewcommand{\arraystretch}{0.9}\resizebox{1.0\textwidth}{!}{\begin{tabular}{llcccccccc}
        \toprule
        \multirow{2}{*}{\textbf{Model}} & \multirow{2}{*}{\textbf{Method}} & \multicolumn{4}{c}{\textbf{Single-Document QA}} & \multicolumn{4}{c}{\textbf{Multi-Document QA}} \\
        \cmidrule(lr){3-6}\cmidrule(lr){7-10}
        & & NarrativeQA & Qasper & MultiFieldQA & \textbf{AVG} & HotpotQA & 2WikiMQA & MuSiQue & \textbf{AVG} \\
        \midrule
        GPT-3.5-Turbo & \multicolumn{1}{c}{-} & 23.6 & 43.3 & 52.3 & 39.8 & 51.6 & 37.7 & 26.9 & 38.7 \\
        \midrule
        \multirow{3}{*}{OL-3B} & Random & \textbf{18.65} & 24.07 & 31.44 & 24.72 & 26.80 & 21.00 & \textbf{11.80} & 19.87 \\
        & ProLong & 17.04 & 25.97 & 32.62 & 25.21 & 27.30 & 23.30 & 10.50 & 20.37 \\
        & LADM & 17.09 & \textbf{28.84} & \textbf{33.97} & \textbf{26.63} & \textbf{29.01} & \textbf{23.71} & 9.15 & \textbf{20.62} \\
        \midrule
        \multirow{3}{*}{L-7B} & Random & 24.06 & \textbf{33.27} & 33.05 & 30.13 & 35.29 & 26.79 & 15.31 & 25.80 \\
        & ProLong & 25.00 & 28.23 & 35.27 & 29.50 & 40.30 & 28.91 & 17.91 & 29.04 \\
        & LADM & \textbf{26.34} & 32.28 & \textbf{38.11} & \textbf{32.24} & \textbf{43.42} & \textbf{31.85} & \textbf{18.03} & \textbf{31.10} \\
        \midrule
        \multirow{3}{*}{M-7B} & Random & 17.30 & 21.39 & 33.56 & 24.08 & 30.67 & \textbf{28.85} & 14.37 & 24.63 \\
        & ProLong & 14.32 & 26.41 & 30.55 & 23.76 & 34.29 & 23.50 & 15.51 & 24.43 \\
        & LADM & \textbf{24.05} & \textbf{34.77} & \textbf{42.72} & \textbf{33.85} & \textbf{39.43} & 28.81 & \textbf{19.04} & \textbf{29.09} \\
        \midrule
        \multirow{3}{*}{L-13B} & Random & 24.07 & 31.31 & 29.32 & 28.23 & 44.88 & 32.36 & 18.82 & 32.02 \\
        & ProLong & \textbf{25.06} & 31.42 & \textbf{30.34} & 28.94 & 44.37 & 34.20 & 21.02 & 33.20 \\
        & LADM & 24.98 & \textbf{33.79} & 28.33 & \textbf{29.03} & \textbf{47.21} & \textbf{36.29} & \textbf{23.09} & \textbf{35.53} \\
        \bottomrule
    \end{tabular}}}
\end{table*}

\begin{table*}[t]\normalsize
    \centering
    {\renewcommand{\arraystretch}{0.9}\scalebox{0.88}{\begin{tabular}{llcccccccc}
        \toprule
        \multirow{2}{*}{\textbf{Model}} & \multirow{2}{*}{\textbf{Method}} & \multicolumn{4}{c}{\textbf{Summarization}} & \multicolumn{3}{c}{\textbf{Code}} & \multirow{2}{*}{\textbf{Overall}} \\
        \cmidrule(lr){3-6}\cmidrule(lr){7-9}
        & & GovReport & QMSum & MultiNews & \textbf{AVG} & LCC & RepoBench-P & \textbf{AVG} \\
        \midrule
        GPT-3.5-Turbo & \multicolumn{1}{c}{-} & 29.5 & 23.4 & 26.7 & 26.5 & 54.7 & 53.6 & 54.1 & 39.8 \\
        \midrule
        \multirow{3}{*}{OL-3B} & Random & 24.33 & 13.83 & 14.01 & 17.39 & 61.21 & 48.69 & 54.95 & 29.23 \\
        & ProLong & \textbf{25.06} & 12.89 & 8.13 & 15.36 & 58.90 & 48.44 & 53.67 & 28.65 \\
        & LADM & 24.04 & \textbf{15.00} & \textbf{15.91} & \textbf{18.32} & \textbf{61.33} & \textbf{49.78} & \textbf{55.56} & \textbf{30.28} \\
        \midrule
        \multirow{3}{*}{L-7B} & Random & 29.56 & 21.26 & 23.66 & 24.83 & 59.32 & 56.51 & 57.92 & 34.67 \\
        & ProLong & 29.54 & 20.49 & 23.35 & 24.46 & 62.78 & 55.30 & 59.04 & 35.51 \\
        & LADM & \textbf{29.77} & \textbf{21.75} & \textbf{26.02} & \textbf{25.85} & \textbf{65.78} & \textbf{58.46} & \textbf{62.12} & \textbf{37.83} \\
        \midrule
        \multirow{3}{*}{M-7B} & Random & 24.68 & 19.93 & 22.91 & 22.51 & 62.95 & 58.74 & 60.85 & 33.02 \\
        & ProLong & 24.43 & 18.92 & \textbf{25.01} & 22.73 & 65.12 & \textbf{58.51} & \textbf{61.82} & 33.18 \\
        & LADM & \textbf{28.38} & \textbf{20.64} & 24.40 & \textbf{24.47} & \textbf{65.41} & 55.74 & 60.58 & \textbf{37.00} \\
        \midrule
        \multirow{3}{*}{L-13B} & Random & \textbf{28.07} & 22.14 & 26.80 & 25.67 & 67.65 & 61.85 & 64.75 & 37.67 \\
        & ProLong & 27.31 & \textbf{22.97} & 26.59 & 25.62 & 67.73 & 61.10 & 64.42 & 38.04 \\
        & LADM & 27.43 & 22.24 & \textbf{27.80} & \textbf{25.82} & \textbf{67.77} & \textbf{62.79} & \textbf{65.28} & \textbf{38.92} \\
        \bottomrule
    \end{tabular}}}
    \caption{\label{longbench}Performance of models trained with different data selection methods on single-document QA, multi-document QA, summarization and code completion from the LongBench dataset.}
    \vspace{-2mm}
\end{table*}
\setlength{\tabcolsep}{4pt}
\begin{table}[!h]
    \centering
    {\renewcommand{\arraystretch}{0.9}\resizebox{1.0\columnwidth}{!}{\begin{tabular}{llcccccc}
        \toprule
        \textbf{Model} & \textbf{Method} & \textbf{Tokens} & \textbf{SD-QA} & \textbf{MD-QA} & \textbf{SUM} & \textbf{CODE} & \textbf{AVG} \\
        \midrule
        \multirow{5}{*}{OL-3B} & \multirow{2}{*}{Random} & 1B & 24.72 & 19.87 & 17.39 & 54.95 & 29.23 \\
        & & 2B & 25.16 & 20.59 & 19.11 & 55.14 & 30.00 \\
        & \multirow{2}{*}{ProLong} & 1B & 25.21 & 20.37 & 15.36 & 53.67 & 28.65\\
        & & 2B & \textbf{27.38} & 18.32 & \textbf{19.20} & 55.26 & 30.04 \\
        & LADM & 1B & 26.63 & \textbf{20.62} & 18.32 & \textbf{55.56} & \textbf{30.28} \\
        \midrule
        \multirow{5}{*}{L-7B} & \multirow{2}{*}{Random} & 1B & 30.13 & 25.80 & 24.83 & 57.92 & 34.67 \\
        & & 2B & 31.91 & 29.12 & 25.18 & 59.86 & 36.52\\
        & \multirow{2}{*}{ProLong} & 1B & 29.50 & 29.04 & 24.46 & 59.04 & 35.51 \\
        & & 2B & 30.10 & 30.58 & 25.84 & 59.52 & 36.51 \\
        & LADM & 1B & \textbf{32.24} & \textbf{31.10} & \textbf{25.85} & \textbf{62.12} & \textbf{37.83} \\
        \midrule
        \multirow{5}{*}{M-7B} & \multirow{2}{*}{Random} & 1B & 24.08 & 24.63 & 22.51 & 60.85 & 33.02 \\
        & & 2B & 29.89 & 26.60 & 23.61 & 58.84 & 34.73 \\
        & \multirow{2}{*}{ProLong} & 1B & 23.76 & 24.43 & 22.73 & 61.82 & 33.18 \\
        & & 2B & 27.41 & 28.85 & 24.29 & \textbf{63.36} & 35.98 \\
        & LADM & 1B & \textbf{33.85} & \textbf{29.09} & \textbf{24.47} & 60.58 & \textbf{37.00} \\
        \midrule
        \multirow{5}{*}{L-13B} & \multirow{2}{*}{Random} & 1B & 28.23 & 32.02 & 25.67 & 64.75 & 37.67 \\
        & & 2B & 27.93 & 34.87 & 25.00 & \textbf{66.13} & 38.48 \\
        & \multirow{2}{*}{ProLong} & 1B & 28.94 & 33.20 & 25.62 & 64.62 & 38.04\\
        & & 2B & \textbf{33.55} & 34.37 & 24.84 & 62.37 & 38.78 \\
        & LADM & 1B & 29.03 & \textbf{35.53} & \textbf{25.82} & 65.28 & \textbf{38.92} \\
        \bottomrule
    \end{tabular}}}
    \caption{\label{eff}Performance comparison of random sampling and our LADM method on the LongBench dataset.}
    \vspace{-8mm}
\end{table}

We compare the performance of our method with baseline methods under different training budgets. As shown in Table \ref{eff}, our method surpasses the baseline even with half of the training tokens. Specifically, LADM achieves slight improvements over OpenLlama-3B and Llama2-13B, exceeding the baseline by 1.31\% for Llama2-7B and 2.27\% for Mistral-7B. Even when the training tokens amount to 3B and 4B for random sampling, our method can still demonstrate superior performance, as illustrated in Appendix \ref{appendixC}. These results highlight that our proposed data selection framework can effectively extract high-quality data with strong contextual dependencies from large-scale pre-training corpora, thus enhancing the long-context modeling capability of LLMs while reducing training costs.

\subsection{Data Selection Efficiency of LADM}

We first analyze the computational complexity of each stage in the LADM framework. Each PFS calculation has a complexity of \scalebox{0.95}{$O(l^2 \cdot d_k)$}. The complexity for AFS is \scalebox{0.95}{$O\left( \frac {N \cdot l^2 \cdot d_k } {d_{AFS}} \right)$} and for the overall CDS is \scalebox{0.95}{$O\left( \frac {N^2 \cdot l^2 \cdot d_k} {d_{CDS} \cdot d_{AFS}}\right) = O\left( \frac{L^2 \cdot d_k}{d_{CDF} \cdot d_{AFS}}\right)$}, where \scalebox{0.85}{$l=128$}, \scalebox{0.85}{$N=256$}, \scalebox{0.85}{$L=32\text{k}$}, and \scalebox{0.85}{$d_{AFS}, d_{CDS}$} are the strides for calculating AFS and CDS. The two parameters decide the number of PFS calculations required and reduce the cost of full attention calculation with complexity \scalebox{0.95}{$O( L^2 \cdot d_k)$}, thus affecting the data selection efficiency.

\begin{table}[!t]
    \centering
    {\renewcommand{\arraystretch}{1.0}\resizebox{1.0\columnwidth}{!}{\begin{tabular}{ccccc}
        \toprule
        \textbf{Method} & \textbf{$d_{AFS}$} & \textbf{$d_{CDS}$} & \textbf{Sec/sample} & \textbf{Correlation} \\
        \midrule
        \multirow{4}{*}{LADM} & 4 & 4 & 2.46 & 1.000 \\
         & 2 & 4 & 2.47 & 0.719 \\
         & 4 & 2 & 3.95 & 0.721 \\
         & 2 & 2 & 3.98 & 0.719 \\
         \midrule
         ProLong & - & - & 2.46 & - \\

        \bottomrule
    \end{tabular}}}
    \caption{\label{hyper}The efficiency and Pearson Correlation under different hyper-parameter settings and methods.}
    \vspace{-5mm}
\end{table}

We conducted experiments on a randomly selected set of 5,000 samples from our pre-training dataset under different hyper-parameter settings. We compare both the computational overhead and the impact on the sample-level CDS. We use the Pearson Correlation Coefficient to measure the consistency of CDS across different configurations. As shown in Table \ref{hyper}, all the coefficient values are greater than 0.7, indicating the consistent results of data selection. For computation efficiency, since we can get all PFS of the current span and its previous spans through multiplication of the query and key matrices, a smaller $d_{AFS}$ does not significantly increase computation overhead. However, smaller $d_{CDS}$ requires calculating additional AFS for new spans, resulting in a notable increase in time overhead. Based on these analyses, we select the setting $d_{AFS}=d_{CDS}=4$ to minimize the computational cost. We also present the data selection efficiency of ProLong in Table \ref{hyper}. With comparable efficiency, LADM can achieve better performance, as shown in Section \ref{section5}, demonstrating the effectiveness and applicability of our method.

\subsection{Ablation Study on LADM}

\setlength{\tabcolsep}{4pt}
\begin{table}[!h]
    \centering
    {\renewcommand{\arraystretch}{0.9}\resizebox{1.0\columnwidth}{!}{\begin{tabular}{lccccc}
        \toprule
        \textbf{Method} & \textbf{SD-QA} & \textbf{MD-QA} & \textbf{SUM} & \textbf{CODE} & \textbf{AVG} \\
        \midrule
        LADM & \textbf{32.24} & \textbf{31.10} & \textbf{25.85} & \textbf{62.12} & \textbf{37.83} \\
        \ \ \ \textit{w/o std} & 30.71 & 30.10 & 25.34 & 61.38 & 36.88 \\
        \ \ \ \textit{w/o length} & 31.60 & 30.90 & 25.38 & 58.08 & 36.49 \\
        \bottomrule
    \end{tabular}}}
    \caption{\label{abl}Ablation study of contextual dependency measurement in LADM.}
    \vspace{-2mm}
\end{table}

To validate the effectiveness of the contextual dependency measurement in LADM, we conduct additional ablation studies on the standard deviation weights $\sigma_{j}$ in Eq. \ref{afs} and the length weights $\frac{j-i}{N}$ and $\frac{j}{N}$ in Eq. \ref{afs} and Eq. \ref{cds}. Experimental results of Llama2-7B on the LongBench dataset are shown in Table \ref{abl}. Higher standard deviation weights indicate more diverse dependency patterns within long-context samples and higher length weights focus more on span pairs with greater distances, making them crucial for long-range contextual dependency measurement. Therefore, we observe a significant performance drop without these weights.

\subsection{Observations and Findings}
We conduct a series of statistical analyses on the sample-level CDS derived by our LADM framework. To mitigate the influence of outliers, we calculate the median scores for each category from the pre-training data.

\begin{figure}[!t]
  \centering
  \includegraphics[width=0.9\columnwidth]{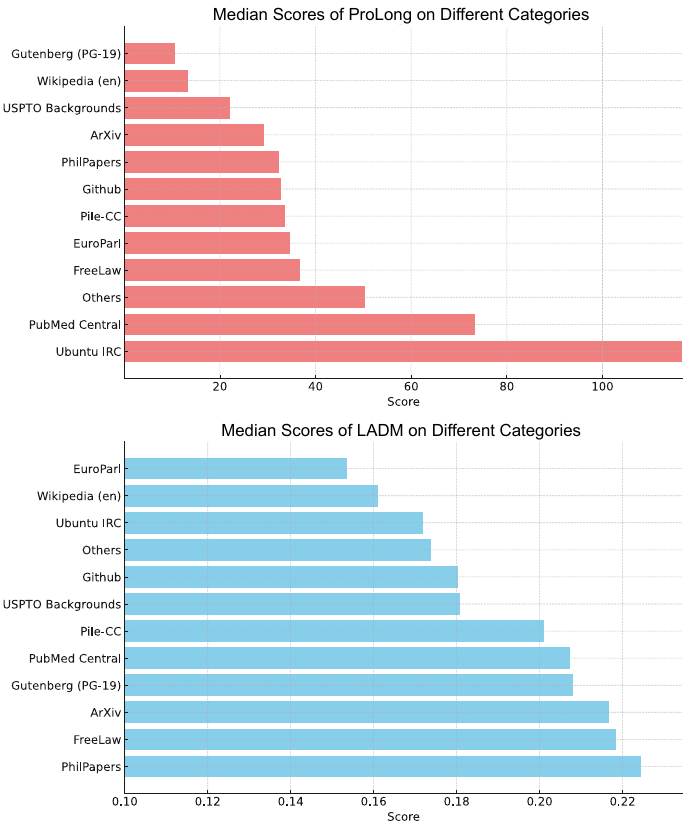}
  \caption{Median scores for various data categories from the Pile dataset under ProLong and LADM framework.}
  \label{cmp_score}
  \vspace{-2mm}
\end{figure}

In Figure \ref{cmp_score}, we show the median scores of ProLong and LADM framework on different categories. The results show that articles (PhilPapers, ArXiv, and PubMed Central), legal documents (FreeLaw), and books (PG-19) tend to receive higher scores in our method. This is likely due to their coherent logic and strong interrelations between paragraphs, indicating complex and varied dependencies within these data types.

Moreover, we analyze the features of data with lower scores. Wikipedia data samples often manifest as concatenated paragraphs that may only share a theme but exhibit weak connectivity between paragraphs. Similarly, the Ubuntu IRC dataset is composed of group chat discussions. These conversations show duplicated patterns and different sections involve different users, time points and topics, resulting in weak contextual coherence. We show samples from different categories in Appendix \ref{appendixD}. 

Compared to our method, ProLong tends to give much higher scores to samples from the Ubuntu IRC dataset containing segments with similar patterns, despite low relevance between them. Moreover, samples from the PG-19 dataset get the lowest score, which indicates that ProLong may struggle to capture the deeper contextual relationships between seemingly unrelated sections within books. These sections may be connected through narrative development, which is not apparent through analysis of isolated segments. This highlights the importance of measuring the dependencies within the full context rather than just assessing the relationship between individual segments.

\section{Conclusion}
This paper introduces LADM, a novel and efficient long-context data selection framework to identify high-quality long-context data with long-range and diverse contextual dependencies from a large-scale, multi-domain pre-training corpus. LADM utilizes the accumulated attention scores over long contexts to quantify the dependencies between spans and aggregate them as a metric for sample-level contextual dependency measurement. The experimental results on various long-context tasks further demonstrate that our method can significantly enhance the long-context modeling capability of LLMs.

\section*{Limitations}
The limitations of our work can be summarized as follows: Firstly, to measure the long-range contextual dependency, we use a tiny model for data selection. This may introduce additional computational overhead. Secondly, we do not conduct experiments on LLMs exceeding 13B parameters, due to the great cost of training long-context LLMs. Thirdly, due to the scarcity of long-context data resources, we only conduct experiments on 32K context length. It is worthwhile to explore ways to construct high-quality data with longer context.

\section*{Acknowledgments}
We thank our colleagues Chen Wang, Chong Li, Zixuan Ren, Yaochen Zhu, and Pu Jian for their insightful and constructive feedback. Furthermore, we thank all reviewers for their detailed reviews and valuable comments. This work is supported by the National Key R\&D Program of China No.2022ZD0160602 and the Strategic Priority Research Program of Chinese Academy of Sciences under Grant No.XDA04080400. This work is also supported by Zhongguancun Academy Project No.20240103.

\bibliography{custom}

\clearpage
\appendix

\section{Pre-training Datasets}
\label{appendixA}

\begin{figure}[h]
  \centering
  \includegraphics[width=0.9\columnwidth]{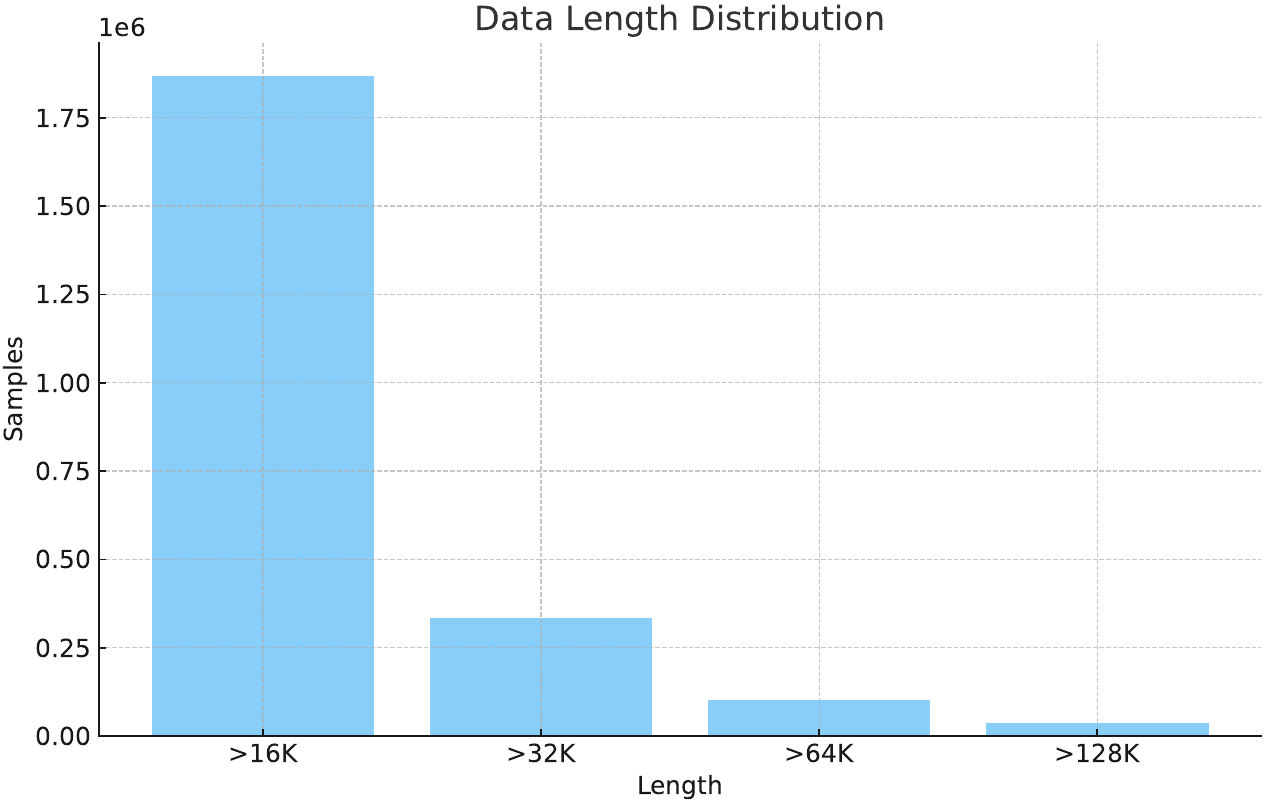}
  \caption{Data length distribution of the Pile dataset.}
  \label{length}
\end{figure}

Figure \ref{length} shows the data length distribution of the Pile dataset. Due to the scarcity of long-context data resources, we conduct experiments on 32K samples without extending to longer data samples. We present the detailed composition of the Pile dataset after removing samples shorter than 32k in Table \ref{app-a}. All subsequent data selection processes are based on this refined dataset.

\begin{table}[h]
    \centering
    {\begin{tabular}{lcc}
        \toprule
        \textbf{Type} & \textbf{Samples} & \textbf{Length} \\
        \midrule
        ArXiv & 216,616 & 49,207 \\
        EuroParl & 17,340 & 71,646 \\
        FreeLaw & 27,663 & 47,264 \\
        Github & 54,844 & 44,336 \\
        Gutenberg (PG-19) & 34,037 & 80,471 \\
        Others & 1,091 & 50,221 \\
        PhilPapers & 3,777 & 69,045 \\
        Pile-CC & 60,066 & 56,367 \\
        PubMed Central & 46,293 & 46,890 \\
        USPTO Backgrounds & 3,762 & 51,124 \\
        Ubuntu IRC & 4,607 & 84,491 \\
        Wikipedia (en) & 4,348 & 52,711 \\
        \bottomrule
    \end{tabular}}
    \caption{\label{app-a}Composition of the pre-training dataset. "Samples" denotes the number of samples and "Length" denotes the average length of samples measured by LlamaTokenizer.}
    \vspace{-2mm}
\end{table}

\begin{figure}[p]
  \centering
  \includegraphics[width=0.9\columnwidth]{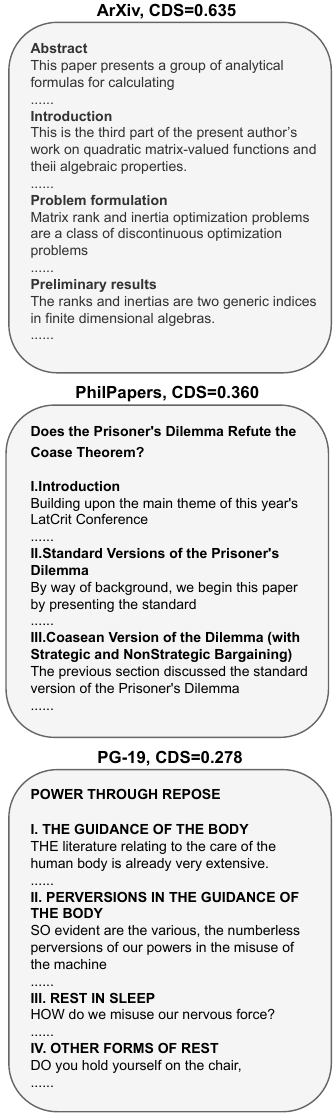}
  \caption{Data samples with high CDS scores from ArXiv, PhilPapers and PG-19.}
  \label{examples_good}
\end{figure}

\section{Training Details}
\label{appendixB} 
We train all models using the language modeling cross-entropy loss and set batch size to 1M tokens and learning rate to $2 \times 10^{-5}$ without weight decay. We employ a linear learning rate scheduler with a warm-up period of 20 steps and use the AdamW optimizer \cite{loshchilov2017decoupled} with $\beta_1 = 0.9$ and $\beta_2 = 0.999$. All models are trained with BF16 precision, Flash Attention 2 \citep{dao2023flashattention2}, HuggingFace Trainer \citep{wolf-etal-2020-transformers}, and Deepspeed engine \citep{rajbhandari2020zero}.

\section{Training Efficiency}
\label{appendixC}
In Table \ref{app-c-new}, we show the performance comparison of LADM with random sampling baseline under different training budgets on Llama2-7B.

\setlength{\tabcolsep}{4pt}
\begin{table}[!h]
    \centering
    \resizebox{1.0\columnwidth}{!}{\begin{tabular}{llcccccc}
        \toprule
        \textbf{Model} & \textbf{Method} & \textbf{Tokens} & \textbf{SD-QA} & \textbf{MD-QA} & \textbf{SUM} & \textbf{CODE} & \textbf{AVG} \\
        \midrule
        \multirow{5}{*}{L-7B} & \multirow{4}{*}{Random} & 1B & 30.13 & 25.80 & 24.83 & 57.92 & 34.67 \\
        & & 2B & 31.91 & 29.12 & 25.18 & 59.86 & 36.52 \\
        & & 3B & 30.28 & 30.83 & 25.33 & 61.91 & 37.09 \\
        & & 4B & 31.81 & 29.86 & 25.37 & 60.37 & 36.85 \\
        & LADM & 1B & \textbf{32.24} & \textbf{31.10} & \textbf{25.85} & \textbf{62.12} & \textbf{37.83} \\
        \bottomrule
    \end{tabular}}
    \caption{\label{app-c-new}Performance comparison between random sampling with additional training tokens and our LADM method on the LongBench dataset.}
    \vspace{-5mm}
\end{table}

\section{Data samples}
\label{appendixD}
As shown in Figure \ref{examples_good}, data samples with high CDS scores are characterized by rich content, coherent logic, and good readability. These samples exhibit interrelations between paragraphs, demonstrating strong long-range contextual dependencies.

\begin{figure}[h]
  \centering
  \includegraphics[width=0.9\columnwidth]{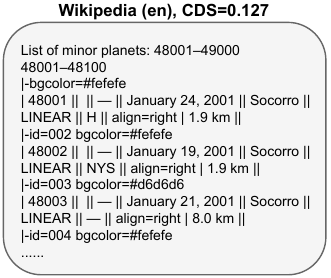}
  \caption{Data sample with low CDS scores from Wikipedia (en).}
  \label{example_wiki}
\end{figure}

\begin{figure}[!h]
  \centering
  \includegraphics[width=0.9\columnwidth]{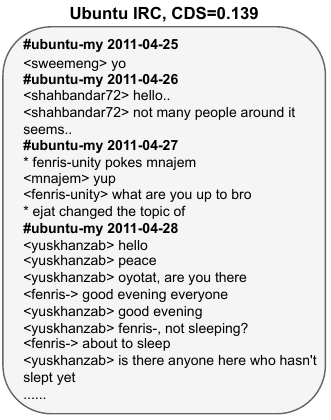}
  \caption{Data sample with low CDS scores from Ubuntu IRC.}
  \label{example_ubuntu}
  \vspace{-2mm}
\end{figure}

\begin{figure}[!h]
  \centering
  \includegraphics[width=0.9\columnwidth]{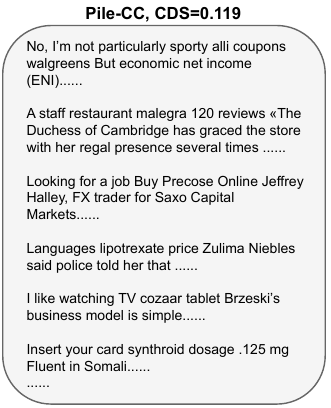}
  \caption{Data sample with low CDS scores from Pile-CC.}
  \label{example_cc}
  \vspace{-4mm}
\end{figure}

\begin{table*}[t]\normalsize
    \centering
    \resizebox{1.0\textwidth}{!}{\begin{tabular}{llccccccccc}
        \toprule
        \multirow{2}{*}{\textbf{Model}} & \multirow{2}{*}{\textbf{Method}} & \multirow{2}{*}{\textbf{LAC-Tokens}} & \multicolumn{4}{c}{\textbf{Single-Document QA}} & \multicolumn{4}{c}{\textbf{Multi-Document QA}} \\
        \cmidrule(lr){4-7}\cmidrule(lr){8-11}
         & & & NarrativeQA & Qasper & MultiFieldQA & \textbf{AVG} & HotpotQA & 2WikiMQA & MuSiQue & \textbf{AVG} \\
        \midrule
        \multirow{4}{*}{L-7B} & Random & - & 24.06 & 33.27 & 33.05 & 30.13 & 35.29 & 26.79 & 15.31 & 25.80 \\
        & ProLong & - & 25.00 & 28.23 & 35.27 & 29.50 & 40.30 & 28.91 & 17.91 & 29.04 \\
        & \multirow{2}{*}{LADM} & 1B & 26.30 & \textbf{34.97} & \textbf{39.56} & \textbf{33.61} & 41.65 & 30.72 & \textbf{19.22} & 30.53 \\
        & & 5B & \textbf{26.34} & 32.28 & 38.11 & 32.24 & \textbf{43.42} & \textbf{31.85} & 18.03 & \textbf{31.10} \\
        \bottomrule
    \end{tabular}}
\end{table*}

\begin{table*}[t]\normalsize
    \centering
    \scalebox{0.83}{\begin{tabular}{lcccccccccc}
        \toprule
        \multirow{2}{*}{\textbf{Model}}& \multirow{2}{*}{\textbf{Method}} & \multirow{2}{*}{\textbf{LAC-Tokens}} & \multicolumn{4}{c}{\textbf{Summarization}} & \multicolumn{3}{c}{\textbf{Code}} & \multirow{2}{*}{\textbf{Overall}} \\
        \cmidrule(lr){4-7}\cmidrule(lr){8-10}
        & & & GovReport & QMSum & MultiNews & \textbf{AVG} & LCC & RepoBench-P & \textbf{AVG} \\
        \midrule
        \multirow{4}{*}{L-7B} & Random & - & 29.56 & 21.26 & 23.66 & 24.83 & 59.32 & 56.51 & 57.92 & 34.67 \\
        & ProLong & - & 29.54 & 20.49 & 23.35 & 24.46 & 62.78 & 55.30 & 59.04 & 35.51 \\
        & \multirow{2}{*}{LADM} & 1B & 29.44 & \textbf{21.97} & 24.89 & 25.43 & 58.81 & 51.65 & 55.23 & 36.20 \\
        & & 5B & \textbf{29.77} & 21.75 & \textbf{26.02} & \textbf{25.85} & \textbf{65.78} & \textbf{58.46} & \textbf{62.12} & \textbf{37.83}\\
        \bottomrule
    \end{tabular}}
    \caption{\label{app-e}Performance comparison of baseline methods and our LADM method using Long Attention Calculator trained with different numbers of tokens (denoted as "LAC-Tokens"). }
\end{table*}
\setlength{\tabcolsep}{6pt}
\begin{table*}[!h]
    \centering
    \begin{tabular}{lccccccc}
        \toprule
        \textbf{Model} & \textbf{Method} & \textbf{MMLU} & \textbf{ARC-c} & \textbf{ARC-e} & \textbf{HellaSwag} & \textbf{TruthfulQA} & \textbf{AVG} \\
        \midrule
        \multirow{4}{*}{OL-3B} & - & 25.26 & 36.52 & 63.80 & 69.91 & 21.30 & 43.36 \\
        & Random & 24.89 & 34.04 & 62.63 & 68.36 & 21.18 & 42.22 \textcolor{green!50!black}{(-1.14)} \\
        & ProLong & 26.11 & 35.84 & 63.68 & 68.46 & 20.56 & 42.93 \textcolor{green!50!black}{(-0.43)} \\
        & LADM & 25.33 & 36.26 & 64.44 & 68.64 & 20.69 & 43.07 \textcolor{green!50!black}{(-0.29)} \\
        \midrule
        \multirow{4}{*}{L-7B} & - & 40.83 & 46.25 & 74.41 & 76.06 & 25.21 & 52.55 \\
        & Random & 37.83 & 44.97 & 74.41 & 75.15 & 26.32 & 51.74 \textcolor{green!50!black}{(-0.81)} \\
        & ProLong & 39.63 & 44.80 & 74.83 & 75.95 & 24.97 & 52.04 \textcolor{green!50!black}{(-0.51)} \\
        & LADM & 40.96 & 45.14 & 74.07 & 76.06 & 26.19 & 52.49 \textcolor{green!50!black}{(-0.06)} \\
        \midrule
        \multirow{4}{*}{M-7B} & - & 59.59 & 53.92 & 79.55 & 81.03 & 28.15 & 60.45 \\
        & Random & 38.17 & 43.09 & 69.40 & 74.65 & 24.72 & 50.01 \textcolor{green!50!black}{(-10.44)} \\
        & ProLong & 44.37 & 49.40 & 77.23 & 78.05 & 25.09 & 54.83 \textcolor{green!50!black}{(-5.62)} \\
        & LADM & 52.76 & 48.12 & 75.34 & 78.51 & 25.46 & 56.04 \textcolor{green!50!black}{(-4.41)} \\
        \midrule
        \multirow{4}{*}{L-13B} & - & 52.09 & 49.40 & 77.61 & 79.37 & 25.34 & 56.76 \\
        & Random & 50.53 & 50.60 & 76.77 & 79.36 & 25.46 & 56.54 \textcolor{green!50!black}{(-0.22)} \\
        & ProLong & 50.61 & 50.00 & 77.31 & 79.44 & 25.95 & 56.66 \textcolor{green!50!black}{(-0.10)} \\
        & LADM & 50.37 & 49.66 & 77.57 & 79.92 & 26.44 & 56.79 \textcolor{red!80!black}{(+0.03)} \\
        \bottomrule
    \end{tabular}
    \caption{\label{app-f}Performance comparison of different models on short-context tasks.}
    \vspace{-2mm}
\end{table*}
We also present samples with low CDS scores. Data from Wikipedia in Figure \ref{example_wiki} exhibits a repetitive structure, primarily consisting of numbers and brief descriptions, lacking detailed information. Chat records Data from Ubuntu IRC in Figure \ref{example_ubuntu} contains a lot of abbreviations and informal expressions. The conversations are composed of content in different time periods, making it difficult to form a coherent information flow. We also observe samples concatenated from completely unrelated data without any logical connection, as shown in Figure \ref{example_cc}. Therefore, these data samples receive lower CDS due to the lack of long-range contextual dependency.

\section{Long Attention Calculator trained with fewer tokens}
\label{appendixE}
In Table \ref{app-e}, we provide the performance of Llama2-7B on the LongBench dataset using Long Attention Calculator trained with different numbers of tokens. The results show that even with fewer training tokens, the Long Attention Calculator can still capture the long-range contextual dependencies and select high-quality data, resulting in the better average performance than baseline methods. This demonstrates the efficiency and robustness of our data selection method.

\section{Performance on short-context tasks}
\label{appendixF}

In Table \ref{app-f}, we present performance of different models on short-context tasks, including MMLU \cite{hendrycks2021measuring}, ARC \cite{clark2018think}, HellaSwag \cite{zellers2019hellaswag} and TruthfulQA \cite{lin2022truthfulqa}. We use the lm-evaluation-harness \cite{eval-harness} for evaluation. Method "-" denotes the original short-context models. For OpenLlama-3B-v2 and Llama2-7B/13B, LADM maintains strong performance on all types of short-text tasks and even shows slight improvement over baseline methods. For Mistral-7B-v0.1, we observe performance decline on all methods, but LADM exhibits the least performance drop compared to other methods.

\end{document}